%% file: main.tex
\documentclass[letterpaper, 10 pt, conference]{ieeeconf}  

\IEEEoverridecommandlockouts                              

\usepackage{booktabs,multirow}
\usepackage{algorithm}
\usepackage[noend]{algpseudocode}
\usepackage[english]{babel}
\usepackage[pdftex]{graphicx}
\usepackage{adjustbox}
\usepackage[colorlinks,linkcolor=red]{hyperref}
\DeclareGraphicsExtensions{.pdf,.jpg,.jpeg,.png}

\usepackage{graphicx} 
\usepackage{color, colortbl}
\usepackage{bm}
\usepackage[caption=false]{subfig}
\usepackage{amsmath} 
\usepackage{amssymb}  
\usepackage{physics}
\usepackage[table]{xcolor}
\usepackage{algcompatible}
\usepackage{multicol}
\usepackage{graphicx}
\usepackage{amsmath}
\usepackage[font=small,labelfont=bf]{caption}
\usepackage{bm}
\usepackage{glossaries}
\usepackage{float}
\usepackage[nodisplayskipstretch]{setspace}
\usepackage{epsfig} 
\usepackage{mathptmx} 
\usepackage{times} 
\newacronym{dofs}{DoFs}{Degrees-of-Freedoms}

\makeatletter
\newcounter {subsubsubsection}[subsubsection]
\renewcommand\thesubsubsubsection{\thesubsubsection .\@arabic\c@subsubsubsection}
\newcommand\subsubsubsection{\@startsection{subsubsubsection}{4}{\z@}%
	{-3.25ex\@plus -1ex \@minus -.2ex}%
	{1.5ex \@plus .2ex}%
	{\normalfont\normalsize\bfseries}}
\newcommand*\l@subsubsubsection{\@dottedtocline{3}{10.0em}{4.1em}}
\newcommand*{\subsubsubsectionmark}[1]{}
\makeatother

\newcommand{\trsp}{{\scriptscriptstyle\top}}

\newcommand{\minus}{\scalebox{0.75}[1.0]{$-$}}

\graphicspath{{./figures/}} 

\DeclareMathOperator*{\argmin}{arg\,min}

\title{\LARGE \bf
	Demonstration-guided Optimal Control\\for Long-term Non-prehensile Planar Manipulation
}

\author{Teng Xue$^{1, 2}$, Hakan Girgin$^{1, 2}$, Teguh Santoso Lembono$^{1,2}$, and Sylvain Calinon$^{1, 2}$ 
\thanks{$^{1}$Idiap Research Institute, Martigny, Switzerland (e-mail: firstname.lastname@idiap.ch)} 
\thanks{$^{2}$École Polytechnique Fédérale de Lausanne (EPFL), Switzerland}%
}

\begin{document}

\maketitle
\thispagestyle{empty}
\pagestyle{empty}

\input{sections/abstracts}
\input{sections/introduction}
\input{sections/related_works}
\input{sections/motion_eq}
\input{sections/methods}
\input{sections/experiments}
\input{sections/conclusion}

\newpage

%
\bibliographystyle{IEEEtran}
\bibliography{main}
%

\end{document}

%% file: sections/abstracts.tex
\begin{abstract}

Long-term non-prehensile planar manipulation is a challenging task for robot planning and feedback control. It is characterized by underactuation, hybrid control, and contact uncertainty. One main difficulty is to determine both the continuous and discrete contact configurations, e.g., contact points and modes, which requires joint logical and geometrical reasoning. To tackle this issue, we propose a demonstration-guided hierarchical optimization framework to achieve offline task and motion planning (TAMP). Our work extends the formulation of the dynamics model of the pusher-slider system to include separation mode with face switching mechanism, and solves a warm-started TAMP problem by exploiting human demonstrations. We show that our approach can cope well with the local minima problems currently present in the state-of-the-art solvers and determine a valid solution to the task. We validate our results in simulation and demonstrate its applicability on a pusher-slider system with a real Franka Emika robot in the presence of external disturbances.\\ Project webpage: \href{https://sites.google.com/view/dg-oc/}{https://sites.google.com/view/dg-oc/}.

\end{abstract}

%% file: sections/introduction.tex
\section{Introduction}
\label{sec:introduction}
With the trend of labor shortage and aging population, robots are required for increasingly  complicated tasks, moving beyond the typical pick-and-place tasks to non-prehensile manipulation, which refers to manipulating without grasping but with mutual interaction. This requires real-time contact identification and adaptation. In this paper, we focus on long-term non-prehensile planar manipulation, which concerns achieving reliable non-prehensile planar manipulation with a long-term horizon, involving joint logistic and geometric planning and feedback control over diverse interaction modes and face switches. For example, to push an object, a prerequisite is to decide how much force should be applied, and which point to push. Moreover, in some cases such as pushing an object with small distance but large orientation, relying on a single fixed face is not feasible. Therefore, a sequence of face switching is required, as well as contact mode schedule resulting from Coulomb friction.

To achieve non-prehensile planar manipulation, four main challenges appear: 
\\$1.$ \textbf{Hybrid Dynamics}. The dynamics of a pusher-slider system depends on the current interaction mode and contact point. In this paper, we consider not only the motion involving various interaction modes, i.e., separation, sticking, sliding up, and sliding down, but also the switching between the contact faces, i.e., left, bottom, right, and up (unlike previous works \cite{hogan_feedback_2020,de2022non} that only work with a single face). 
\\$2.$ \textbf{Underactuation}. The contact force between the pusher and the slider is constrained within a motion cone, making it impossible to exert arbitrary acceleration on the object to achieve omnidirectional movement.
\\$3. $ \textbf{Long-horizon TAMP}. Due to the characteristics of hybrid dynamics and underactuation, it is important to reason over long horizon using both logic and geometric descriptors, where the logic variables relate to contact modes and faces and the geometric variables relate to contact points, slider states, switching points, and control inputs.
\\$4. $ \textbf{Contact Uncertainty}. Arising from the frictional contact interactions between the pusher and the slider, as well as between the slider and the table, the contact is hard to be modeled precisely, therefore a controller enabling online contact adaptation is required. 

\begin{figure}[t]
	\centering
	\includegraphics[width=0.75\columnwidth]{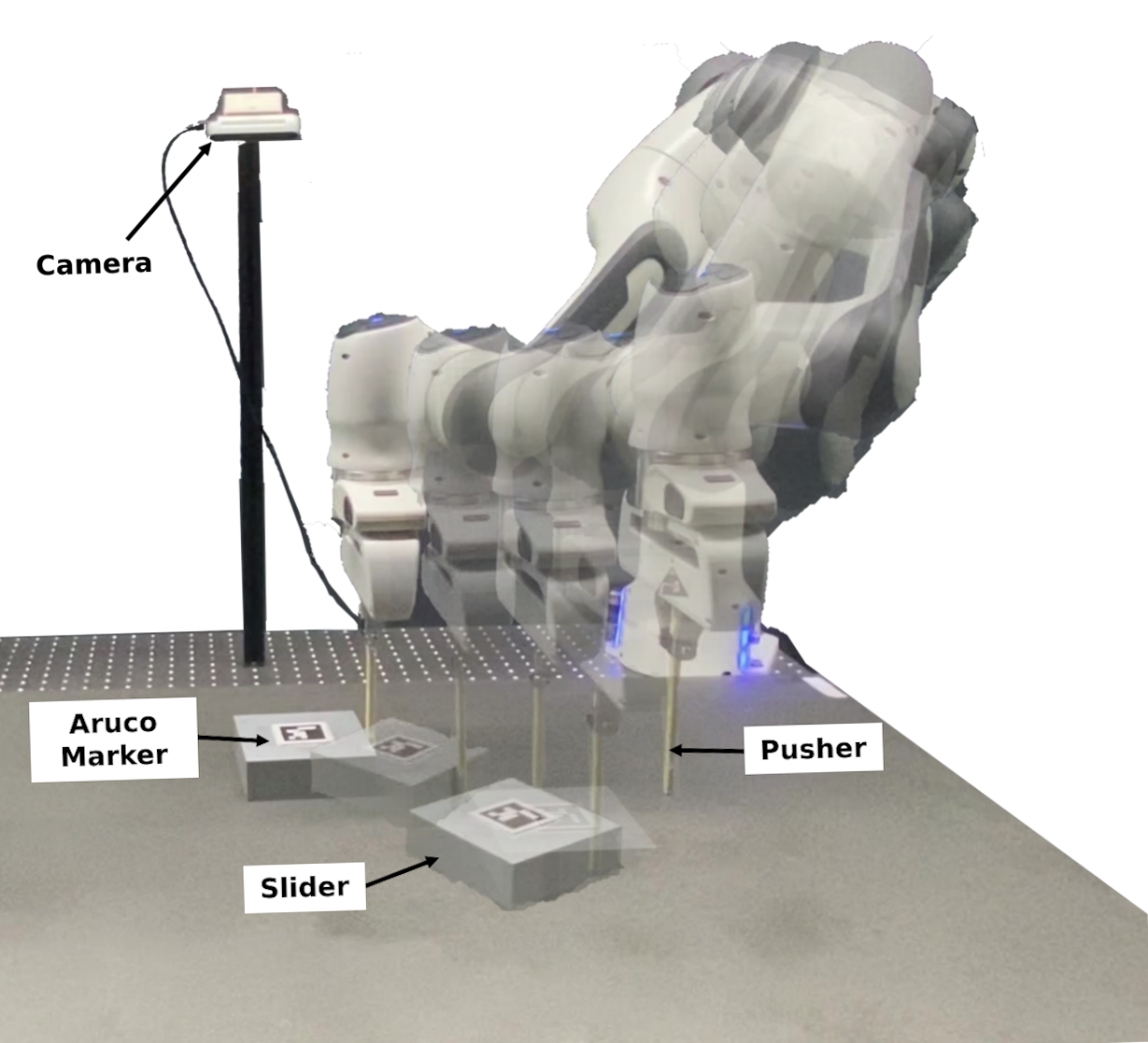}
	\caption{System setup where the robot pushes an object with changes of contact points.}
	\label{fig:real_setup}
	\vspace{-0.5cm}
\end{figure}

The existence of both continuous and discrete variables characterizes the hybrid nature of contact, making it challenging for gradient-based optimization algorithms. Although optimal control has demonstrated good results for contact-rich tasks through techniques such as contact smoothing \cite{pang2022global} and the inclusion of complementary constraints \cite{de2022non}, it is still susceptible to poor local minima. On the other hand, sampling-based methods like search tree \cite{doshi2020hybrid} have the ability to perform long-term global planning, but can struggle with the vast number of potential contact modes. Human decision-making, conversely, excels at discrete decisions. Integrating human demonstration into the hybrid optimization framework could greatly enhance its convergence towards near-optimal solutions.


The main contribution of our paper is a hybrid framework that integrates optimization and learning from human demonstrations to overcome the issue of getting trapped in poor local optima, which is a common problem in many state-of-the-art solvers. Specifically, we demonstrate the effectiveness of our approach in a challenging long-term non-prehensile manipulation task that involves multiple interaction modes and face-switching cases. We collect demonstrations through an interface\footnote{\url{https://sites.google.com/view/dg-oc/}}, where humans are required to guide the robot to push the block toward the target. The discrete decision variables are represented implicitly by some continuous variables. We then use the collected demonstration data as soft constraints in the cost function for optimization, enabling efficient exploration of the solution space that includes both logic and geometric variables. Additionally, we propose a real-time feedback controller that accounts for model mismatch and contact uncertainty when interacting with the physical world. It is worth noting that our approach prioritizes motion planning in the task space and then utilizes operational space control \cite{siciliano2008springer} to compute joint torques based on end-effector motion.

%% file: sections/related_works.tex
\section{Related Work}
\label{sec:related work}
In this section, we discuss some related work on dynamics modeling, TAMP, and robot learning from demonstration.

Non-prehensile manipulation has been widely studied as a challenging task for model-based planning and control \cite{mason1999progress}, with the pusher-slider system as one of the most prominent example. Under the assumption of quasi-static interaction \cite{goyal1989limit} and  ellipsoidal limit surface \cite{lee1991fixture}, Lynch \emph{et al.} put forward a set of analytical methods to compute velocities based on the motion cone \cite{lynch1992manipulation}, which was previously studied in \cite{mason1986mechanics} to distinguish different sticking and sliding behaviors with a friction cone. Recently, Hogan \emph{et al.} reformulated the motion equation as a piecewise function to describe hybrid dynamics \cite{hogan_feedback_2020}. In \cite{hogan2018reactive}, they instead expressed the limit surface in convex quadratic form to simplify the previous equation as a more general form. Nevertheless, the motion equations mentioned above are all assuming that the contact face is fixed, which is too constrained for long-term manipulation, since the pusher will lose the chance to adjust the contact point and face during pushing. Therefore, an extended dynamics model is required to provide the pusher with more flexible choices on the contact configuration.

A hybrid framework with planning and feedback control is important to achieve long-term non-prehensile planar manipulation. Optimally unifying hybrid interaction modes and geometric variables is still an open challenge. For instance, in \cite{hogan_feedback_2020}, a mixed integer programming (MIP) was introduced into a model predictive controller (MPC) to incorporate the selection of interaction modes, i.e., sticking and sliding, into a common optimization framework. Due to the non-convex nature of integer variables, the computation can be time-consuming and not feasible for high-frequency feedback controller. An offline 3-layer mode sequence classifier was later proposed in \cite{hogan2018reactive}. A higher number of discrete modes and contact faces would require the training dataset to increase exponentially, leading to more time-consuming data generation process. Moreover, the works mentioned above are both based on short-horizon MPC, which is capable of online feedback control but shortsighted for long-term planning, which typically does not allow for separation mode and face-switching cases. Alternatively, a mathematical program with complementarity constraints (MPCC) was proposed in \cite{de2022non}, showing faster convergence for planning and recovery from external disturbance, but still susceptible to poor local optima due to the local smoothing and relaxation. 

To cope with this issue, sampling-based methods have been used to build a search tree to optimize the symbolic nodes with geometric variables \cite{simeonov2021long, toussaint2015logic, migimatsu2020object, fikes1971strips}. In the field of non-prehensile manipulation, a search tree with predefined depth was used in \cite{doshi2020hybrid}, corresponding to the number of hybrid switches to optimize over a fixed contact modes sequence. A hybrid Differential Dynamic Programming (DDP) algorithm was simultaneously employed for geometric reasoning. However, the method was validated only with a simplified setup, without considering sliding and separation mode, and without considering optimal viapoints for face switching. Hereby, with additional logic variables, the nodes in the search tree would exponentially grow, yielding exponential computation. Moreover, although expensive sampling computation can be resolved by multiple threads with a powerful machine, there is no guarantee that a feasible solution can be solved in a finite time. 

In contrast to such long-term TAMP formulations, humans showcase an impressive agility at exploiting (sub)optimal strategies, especially in terms of discrete variables, by relying on past experience. In this direction, Learning from Demonstration (LfD) has been studied in robotics for a long time and has proven to be an efficient way to solve motion planning challenges \cite{Billard16chapter, Calinon19chapter}. LfD aims to extract motion features from only few demonstrations and then generalize the learned tasks to new situations. Many algorithms have been proposed to encode human demonstrations, such as Dynamic Movement Primitive (DMP) \cite{ijspeert2013dynamical}, Gaussian Mixture Regression (GMR) \cite{Calinon19chapter} and task-parameterized probabilistic models \cite{calinon2018robot}. Among these techniques, K-nearest neighbor (k-NN) \cite{peterson2009k} is the simplest, yet remaining a powerful tool for discrete data classification \cite{cunningham2021k}. 

%% file: sections/motion_eq.tex
\renewcommand{\arraystretch}{1.}
\begin{table*}[htbp]
	\centering    
	\caption{Constraints of different interaction modes}
	\setlength{\tabcolsep}{18pt}
	\begin{footnotesize}
		\begin{tabular}{|c|c|c|c|}
			\toprule
			\textbf{Sticking} &\textbf{Sliding Up} &\textbf{Sliding Down} &\textbf{Separation} \\
			\midrule
			$\{\bm{v}_f\in \Omega\}  \cap \{\bm{q}_{p_f} \in \mathcal{\psi}\}$
			&$\{\Omega < \bm{v}_f < \mathcal{\phi}\}\cap \{\bm{q}_{p_f} \in \mathcal{\psi}\}$
			&$\{\mathcal{\phi} < \bm{v}_f < \Omega\} \cap \{\bm{q}_{p_f} \in \mathcal{\psi}\}$
			&$\{\bm{v}_f \notin \mathcal{\phi}\} \cup  \{\bm{q}_{p_f} \notin \mathcal{\psi}\}$\\
			\midrule
			$\begin{aligned}
				&\frac{v_{tf}}{v_{nf}} \leq \gamma_{up}, \quad
				\frac{v_{tf}}{v_{nf}} \geq \gamma_{dn}, \quad v_{nf} > 0, \\ &\left |p_x\right | = r_s + r_p, \quad
				\left |p_y\right | \leq r_s.
			\end{aligned}$
			&$\begin{aligned}
				&\frac{v_{tf}}{v_{nf}} > \gamma_{up}, \quad
				v_{nf} > 0,\\
				&\left |p_x\right | = r_s + r_p, \quad \left |p_y\right | \leq r_s.
			\end{aligned}$
			& $\begin{aligned}
				&\frac{v_{tf}}{v_{nf}} < \gamma_{dn}, \quad
				v_{nf} > 0,\\
				&\left |p_x\right | = r_s + r_p, \quad \left |p_y\right | \leq r_s.
			\end{aligned}$ 
			& $\begin{aligned}
				&\vee v_{nf} < 0, \\
				&\vee \left |p_x\right | > r_s+r_p, \\
				&\vee \left |p_y\right | > r_s.
			\end{aligned}$\\
			\bottomrule
		\end{tabular}
	\end{footnotesize}
	\label{tab:motion_cone}
	\vspace{-0.5cm}
\end{table*}

\section{Dynamical System}
\label{sec:method1}

In this section, the hybrid dynamics of the pusher-slider system is described. We introduce the kinematics model (Sec. \ref{sec:kinematics}) first, and then extend motion cone (Sec. \ref{sec:motion cone}) and motion equation (Sec. \ref{sec:motion equation}) by including separation mode and face-switching mechanism.  

\subsection{Kinematics}
\label{sec:kinematics}
\begin{figure}[h]
	\centering
	\includegraphics[width=0.5\columnwidth]{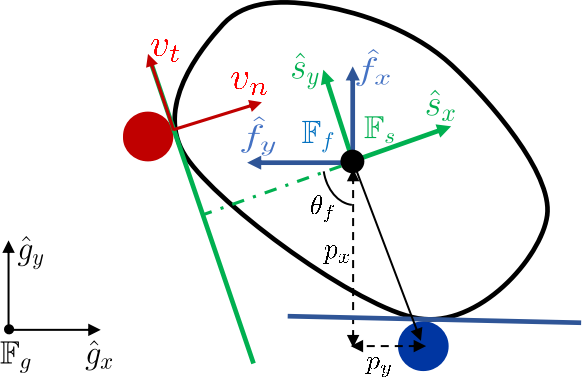}
	\caption{Kinematics of pusher-slider system allowing face switching.}
	\label{fig:kinematics}
	\vspace{-0.3cm}
\end{figure}

Fig. \ref{fig:kinematics} shows the kinematics of the pusher-slider system. The red and blue circles represent the pusher at the initial face and the face after switching. The pose of the slider is defined as $\bm{q}_s = [x \ y \ \theta]^\trsp$ w.r.t.~the global frame $\mathbb{F}_g$, where $x$ and $y$ are the Cartesian coordinates of the center of mass, and $\theta$ is the rotation angle around the vertical axis. The contact position between the pusher and the slider is described as $\bm{q}_{p_f} = [p_x \ p_y]^\trsp$ w.r.t.~the current slider frame $\mathbb{F}_f$ after face switching, while $\bm{q}_{p_s} = \bm{R}_{\theta_f}\bm{q}_{p_f}$ is the expression of $\bm{q}_{p_f}$ in the initial slider frame $\mathbb{F}_s$, and $\theta_f$ is the rotation angle from $\mathbb{F}_s$ to  $\mathbb{F}_f$. The input of this system is defined in $\mathbb{F}_s$ as the acceleration of the pusher $\bm{u} = [\dot{v_n} \ \dot{v_t}]^\trsp$, while $\bm{v} = [v_n \ v_t]^\trsp$ is the pusher velocity. $\bm{v}_f = \bm{R}_{\theta_f}^\trsp \bm{v} = [v_{nf} \ v_{tf}]^\trsp$ denotes $\bm{v}$ in frame $\mathbb{F}_f$.
\vspace{-0.3cm}

\subsection{Generalized Motion Cone}
\label{sec:motion cone}

Based on the quasi-static approximation and ellipsoidal limit surface assumption \cite{hogan_feedback_2020,goyal1989limit,lee1991fixture,mason1986mechanics}, a motion cone is introduced to determine the contact mode given by the current pusher velocity and contact position. The two boundaries of the motion cone are given as $\bm{v_{up}} = 1\bm{f_x} + \gamma_{up}\bm{f_y}$ and $\bm{v_{dn}} = 1\bm{f_x} + \gamma_{dn}\bm{f_y}$, resolved in the current slider frame $\mathbb{F}_f$, with 
\begin{equation}
	\begin{small}
		\gamma_{up} = \frac{\mu_p c^2-p_x p_y+\mu_p {p_x}^2}{c^2+{p_y}^2-\mu_p p_x p_y},
	\end{small}
	\label{eq:gamma_up}
\end{equation}
\begin{equation}
	\begin{small}
		\gamma_{dn} = \frac{-\mu_p c^2-p_x p_y-\mu_p {p_x}^2}{c^2+{p_y}^2+\mu_p p_x p_y},
	\end{small}
	\label{eq:gamma_dn}
\end{equation}
where $\mu_p$ is the friction coefficient between the pusher and the slider, and $c$ is the parameter connecting applied force and the resulting velocity. More details are introduced in \cite{hogan_feedback_2020}.

In the previous work \cite{hogan_feedback_2020, lynch1992manipulation}, the pusher is assumed to remain on one fixed contact face during the execution. We would like to enable more complex pushing strategies by involving face switching. For that, we introduce another interaction mode, which we call separation mode. Table \ref{tab:motion_cone} lists the relationship between state constraints and interaction modes, where $\Omega$ is the set within the boundaries of the motion cone, $\mathcal{\phi}$ is the space where the pusher goes towards the slider, and $\mathcal{\psi}$ is the set where the pusher keeps touching with the slider. $r_s$ and $ r_p$ are the half length of the slider and the radius of the pusher, respectively.

%
%

\subsection{Generalized Motion Equation}
\label{sec:motion equation}
We build the generalized motion equation on the basis of \cite{lynch1992manipulation} and \cite{hogan_feedback_2020}, by including separation mode and contact face switching, namely
\begin{equation}
	\bm{\dot{x}}=\left\{
	\begin{aligned}
		& \bm{g}_1(\bm{x}, \bm{u}), \quad \text{if} \quad \textbf{Sticking}, \\
		& \bm{g}_2(\bm{x}, \bm{u}), \quad \text{if} \quad \textbf{Sliding Up}, \\
		& \bm{g}_3(\bm{x}, \bm{u}), \quad \text{if} \quad \textbf{Sliding Down}, \\
		& \bm{g}_4(\bm{x}, \bm{u}), \quad \text{if} \quad \textbf{Separation},
	\end{aligned}
	\right.
	\label{eq:motion_eq}		
\end{equation}
where $\bm{x} = [\bm{q}_s^\trsp \ \bm{q}_{p_s}^\trsp \ \bm{v}^\trsp]^\trsp$, and
\begin{small}
	\begin{equation*}
		\bm{g}_j(\bm{x}, \bm{u}) = \left[ \begin{matrix} \left[
			\begin{matrix}
				\bm{R}_{\theta_f} \bm{R}_\theta \bm{Q} \bm{P}_j \\
				\bm{b}_j \\
				\bm{R}_{\theta_f} [\bm{d}_j \quad \bm{c}_j]^\trsp
			\end{matrix} \right]
			\bm{R}_{\theta_f}^\trsp \bm{v} \\
			\bm{u}
		\end{matrix}  \right], \quad
		\bm{R}_\theta = \left[
		\begin{matrix}
			\cos\theta & -\sin\theta \\
			\sin\theta & \cos\theta
		\end{matrix}
		\right], 
	\end{equation*}
	\begin{equation*}
		\bm{Q} = \frac{1}{c^2 + p_x^2 + p_y^2} \left[
		\begin{matrix}
			c^2 + {p_x}^2 & p_x p_y \\
			p_x p_y & c^2 + p_y ^2
		\end{matrix}
		\right],
	\end{equation*}
	\begin{equation*}
		\bm{b}_1 = \Big[\frac{-p_y}{c^2+p_x^2+p_y^2} \quad  \frac{p_x}{c^2+p_x^2+p_y^2}\Big], \\
		\bm{b}_2 = \Big[\frac{-p_y+\gamma_{up}p_x}{c^2+p_x^2+p_y^2} \quad 0\Big],
	\end{equation*}
	\begin{equation*}
		\bm{b}_3 = \Big[\frac{-p_y+\gamma_{dn}p_x}{c^2+p_x^2+p_y^2} \quad 0\Big], \quad
		\bm{b}_4 = [0 \quad 0],
	\end{equation*}
	\begin{equation*}
		\begin{aligned}
			\bm{c}_1 = [0 \quad 0], \quad
			\bm{c}_2 = [-\gamma_{up} \quad 1], 
			\bm{c}_3 = [-\gamma_{dn} \quad 1],
			\bm{c}_4 = [0 \quad 1],
		\end{aligned}
	\end{equation*}
	\begin{equation*}
		\bm{d}_1 = [0 \quad 0], \quad
		\bm{d}_2 = [0 \quad 0], \quad
		\bm{d}_3 = [0 \quad 0], \quad
		\bm{d}_4 = [1 \quad 0], \quad
	\end{equation*}
	\begin{equation*}
		\bm{P}_1 = \bm{I}_{2\times2}, \quad
		\bm{P}_2 = \left[
		\begin{matrix}
			1 & 0 \\
			\gamma_{up} & 0
		\end{matrix}
		\right], 
		\bm{P}_3 = \left[
		\begin{matrix}
			1 & 0 \\
			\gamma_{dn} & 0
		\end{matrix}
		\right], \quad
		\bm{P}_4 = \bm{0}_{2\times2}.
	\end{equation*}
\end{small}

\noindent where $j = 1, 2, 3, 4$ corresponds to sticking, sliding up, sliding down, and separation mode, respectively.

%% file: sections/methods.tex
\section{Methods}
\label{sec:method}
In this section, we first formulate the long-term non-prehensile planar pushing task as an optimal control problem (Sec. \ref{sec:ocp_formu}), and then we propose three DDP-related methods by modifying the initialization and cost function, namely Demonstration-started DDP (Sec. \ref{sec:DS_DDP}), Demonstration-penalized DDP (Sec. \ref{sec:ddp_dem}) and Warm-starting DDP (Sec. \ref{sec:warm-start DDP}). The combination of Demonstration-penalized DDP and Warm-starting DDP leads to a demonstration-guided hierarchical optimization framework. Finally, we describe the online tracking controller (Sec. \ref{sec:adap_dist}).

\subsection{Problem formulation}
\label{sec:ocp_formu}

An optimal control problem (OCP) can be described as
\vspace{-0.3cm}
\begin{align} \label{eq:basic-cost}
	&\min_{\bm{u}_t} \quad c_T(\bm{x}_T) + \sum_{t=0}^{T-1} c_t(\bm{x}_t,\bm{u}_t) ,\\
	&\begin{array}{r@{\quad}r@{}l@{\quad}l}
		\text{s.t.}  \quad \bm{x}_{t+1} = \bm{f}(\bm{x}_t, \bm{u}_t),
		\label{eq:dynamics}
	\end{array}
\end{align}
\noindent where~\eqref{eq:basic-cost} is the cost function and~\eqref{eq:dynamics} is the dynamic equation. Due to the nonlinearity and high degrees of freedom in robotics, numerical optimization is commonly used to solve such problems. This paper employs DDP (Differential Dynamic Programming) \cite{mayne1966second}, which has been demonstrated to be effective in previous studies \cite{doshi2020hybrid,tassa2014control}. Furthermore, DDP provides a local feedback mechanism that enhances the robustness of the controller, as demonstrated in \cite{Girgin22}.

After minimizing the cost-to-go function w.r.t.~$\Delta\bm{u}_t$, a local stabilizing controller can be obtained as
\begin{equation}
	\bm{u}_t = \hat{\bm{u}}_t + \bm{K}_t \, (\bm{x}_t - \hat{\bm{x}}_t),
	\label{eq:iLQRrecursiveController}
\end{equation}
where $\bm{K}_t$ is the feedback gain, and $\hat{\bm{u}}_t$ is the feedforward term.

Given \eqref{eq:basic-cost} is a non-convex problem, DDP solves it by optimizing around the current solution iteratively. The convergence is very sensitive to the initial guess, which means that the algorithm can converge to poor local optima if the initial guess is far away from the optimal solution.
\subsection{Demonstration-started DDP (DS-DDP)}
\label{sec:DS_DDP}
Demonstrations can be used for OCP initialization to address the poor local optima problem \cite{lembono2020memory, mansard2018using}. Here, we also use human demonstration for initialization. The demonstrations are designed as continuous variables, which can implicitly express the discrete variables instead of explicitly specifying the mode sequence as in previous work \cite{hogan_feedback_2020, de2022non,  hogan2018reactive}. In this way, the joint logic and geometric optimization problem can be converted to a continuous optimization problem that can be solved efficiently.

The collected human demonstrations are denoted as $[\bm{\widetilde{q}}_s, \bm{\widetilde{q}}_{p_s}, \bm{\widetilde{v}}, \bm{\widetilde{u}}]$, with $\bm{\widetilde{q}}_s = [\bm{\widetilde{q}_{s_0}}, \bm{\widetilde{q}_{s_1}}, \cdots,\bm{\widetilde{q}_{s_{T}}}]$, $\bm{\widetilde{q}}_{p_s} = [\bm{\widetilde{q}_{p_{s_0}}}, \bm{\widetilde{q}_{p_{s_1}}}, \cdots, \bm{\widetilde{q}_{p_{s_T}}}]$, where $\bm{\widetilde{q}}_{s_t} \in \mathbb{R}^{3}$ and $\bm{\widetilde{q}}_{p_{s_t}} \in \mathbb{R}^{2}$ represent the state of the slider and the pusher at timestep $t$, respectively. $\bm{\widetilde{v}} = [\bm{\widetilde{v}}_0, \bm{\widetilde{v}}_1, \cdots,\bm{\widetilde{v}}_{T-1}]  \in \mathbb{R}^{2}$ and $\bm{\widetilde{u}} = [\bm{\widetilde{u}}_0, \bm{\widetilde{u}}_1, \cdots,\bm{\widetilde{u}}_{T-1}]  \in \mathbb{R}^{2}$ denote the velocity and acceleration at each timestep.

Given a target $\bm{q}_s^*$, we use k-NN to select the index of $j^*$ with the closest demonstration $\bm{\widetilde{q}}_{s_{T}}$ to the target in the task space by evaluating
\begin{equation}
	\begin{small} 
		j^* = \argmin_{j \in \mathbb{S}} \quad \text{dist} (\bm{\widetilde{q}}^j_{s_{T}}, \bm{q}_s^*),
	\end{small}
	\label{eq:k-NN}
\end{equation}
where $\mathbb{S} = \{j: j \in \{0, 1, \cdots, n_d-1 \} \}$, $n_d$ is the number of demonstrations, and
\begin{equation} 
	\begin{small}
		\text{dist} (\bm{x}, \bm{y}) = 	\Bigg(\sum_{r=1}^d |x_r - y_r|^p\Bigg)^{1/p},
	\end{small}
	\label{eq:dist}
\end{equation}
where d is the dimension of slider state. $p$ is set as 2.

The cost function of DS-DDP is defined as
\begin{equation} 
	\begin{gathered}
		c_{1} = c_{re} + c_{rg} + c_{bd},
	\end{gathered}
	\label{eq:cost_DS-DDP}
\end{equation}
with
\begin{equation*}
	\begin{footnotesize}
		\begin{aligned}
			&c_{re} = (\bm{\mu}_T - \bm{x}_T)^\trsp \bm{Q}_T (\bm{\mu}_T - \bm{x}_T), \quad c_{rg} =
			\sum_{t=0}^{T-1} \bm{u}_t^\trsp \bm{R} \bm{u}_t,\\
			&c_{bd} = \sum_{t=0}^{T-1} \bm{f}^\text{cut}(\bm{u}_t, \bm{u}_l)^\trsp \bm{Q}_f \bm{f}^\text{cut}(\bm{u}_t, \bm{u}_l), 
		\end{aligned}
	\end{footnotesize}
\end{equation*}
where $c_{re}$ is the reaching cost, and $c_{rg}$, $c_{bd}$ are the regularizer and boundary penalizer of control commands. $\bm{u}_l$ is the predefined bounding box of $\bm{u}$. $\bm{f}^\text{cut}$ is a soft-thresholding function. The initial guess ${\bm{u}^0} = \bm{\widetilde{u}}$ is drawn from human demonstrations directly.

Although this method seems like an effective warm-starting method, it is restricted by the number of acquired demonstrations.

\subsection{Demonstration-penalized DDP (DP-DDP)}
\label{sec:ddp_dem}
To alleviate the problem mentioned above, we propose to use demonstration as soft constraints. It is achieved by designing the cost function as
\begin{equation} 
	c_{2} =  c_{re} + c_{rg} + c_{bd} + c_{sw} + c_{ve} + c_{ac},
	\label{eq:cost_demGuided}
\end{equation}
with 
\begin{equation*}
\begin{footnotesize}
\begin{aligned}
	&c_{sw} = \sum_{n=t_0}^{t_{N-1}}(\bm{\mu}_n - \bm{x}_n)^\trsp \bm{Q}_n (\bm{\mu}_n - \bm{x}_n),\\
        &c_{ve} = \sum_{t=0}^{T-1} (\bm{\widetilde{v}}_t - \bm{v}_t)^\trsp \bm{R}_{dv} (\bm{\widetilde{v}}_t - \bm{v}_t),\\
	&c_{ac} = \sum_{t=0}^{T-1} (\bm{\widetilde{u}}_t - \bm{u}_t)^\trsp \bm{R}_{du} (\bm{\widetilde{u}}_t - \bm{u}_t),
\end{aligned}
\end{footnotesize}
\end{equation*}
where $c_{sw}$, $c_{ve}$ and $c_{ac}$ are designed to follow the demonstrated face switching strategy, pusher velocity and pusher acceleration. $n=[t_0, \cdots, t_{N-1}]$ is the timestep when the contact face switches, $\bm{\mu} = [\bm{\widetilde{q}}_s^\trsp \ \bm{\widetilde{q}}_{p_s}^\trsp \ \bm{\widetilde{v}}^\trsp]^\trsp$ is the state of the selected demonstration, and $\bm{\widetilde{v}}_t$, $\bm{\widetilde{u}}_t$ are the demonstrated velocity and acceleration at timestep $t$.

\subsection{Warm-starting DDP (WS-DDP)}
\label{sec:warm-start DDP}

Demonstration-penalized DDP shows good performance and can avoid poor local optima, but in order to further improve its convergence properties, we propose a hierarchical optimization framework, where the solution of DP-DDP is used to initialize another DDP problem that we call Warm-starting DDP (WS-DDP).

The cost function of WS-DDP is as same as DS-DDP, allowing it to explore much freely towards the final target. The initial guess ${\bm{u}^0} = \bm{u}^*_{DP}$ is the result of previous DP-DDP.

\subsection{Adaptation to disturbance}
\label{sec:adap_dist}
 Typically, the feedback controller \eqref{eq:iLQRrecursiveController} is used to generate control commands at each timestep based on the current state to stabilize the motion along the nominal trajectory. However, the friction and other unmodeled nonlinearities might cause undesired behaviors, especially when the error between the current state and the planned solution $\norm{\bm{x}_t-\bm{\hat{x}}_t}^2$ is too big. To alleviate this issue, inspired by \cite{Girgin22}, we propose to use an error filtering method based on a trust region defined as a ball of radius $r$ as $\mathcal{B}_t(r)=\{\bm{x}\in \mathbb{R}^n|\norm{\bm{x}-\bm{\hat{x}}_t} < r \}$. By denoting the actual robot state as $\bm{x}_t^0$, we filter the $\bm{x}_t$ in \eqref{eq:iLQRrecursiveController} as follows: if $\bm{x}_t^0 \in \mathcal{B}_t(r)$, then we take $\bm{x}_t=\bm{\hat{x}}_t$, else we take $\bm{x}_t=\bm{x}_t^0$. This means that if the error between the actual robot state and the planned state is small, we can use the feedforward terms of the controller directly, and if it is big, then we regenerate the trajectory using the feedback controller gains. This allows us to fully exploit the feedback and feedforward gains computed offline during the online execution of the task avoiding expensive recomputations at each timestep.

%% file: sections/experiments.tex
\section{Experiments}
\label{sec:experiments}
We evaluate in this section the proposed offline programming method (Sec. \ref{sec:ex-offline programming}) and the online tracking controller (Sec. \ref{sec:ex_online tracking}). 

\subsection{Offline Programming}
\label{sec:ex-offline programming}
In this work, the task space is the horizontal plane on the table, and it is limited as $\mathcal{T} = \{[x, y, \theta]: x \in [\minus 25\text{cm}, 25\text{cm}], y \in [\minus 25\text{cm}, 25\text{cm}], \theta \in [\minus \pi, \pi]\}$. We collected 3 representative demonstrations, which are $[15\text{cm}, -10\text{cm}, \minus \pi/2], [0, \minus 20\text{cm}, \pi/2], [15\text{cm}, -15\text{cm}, \pi/2]$, corresponding to $N_s=0$, $N_s=1$ and $N_s=2$, respectively, where $N_s$ is the number of face switches during the demonstration. The initial state is defined as $[0,\;0,\;0,\;\alpha p_x,\;0,\;0,\;0]^\trsp$, where $\alpha = 1.3$, corresponding to the separation mode in Sec. \ref{sec:motion cone}, allowing the pusher to select the contact point at the beginning. The cost function gains are set to $\bm{Q}_T = 10^6\times\text{diag}\{1, 1, 1, 10^{6p-5}, 10^{1-6p}, 10^{-3}, 10^{-3}\}$,  $\bm{Q}_n = 10^6\times\text{diag}\{10^{-3},10^{-3},10^{-3}, p, (1-p), 0, 0\}$, where $p=1$ when $\theta_f=0$ or $\theta_f=\pi$, otherwise $p=0$. $\bm{R}=\bm{K}=\text{diag}\{1,1\}$, $\bm{R}_{du}=\bm{R}_{dv}=\text{diag}\{100, 100\}$.

We first randomly selected 100 targets from the task space $\mathcal{T}$ to test our proposed Demonstration-guided Hierarchical Optimization framework, along with two benchmarks: Zero-started DDP (ZS-DDP), namely without using demonstration, and Demonstration-started DDP (DS-DDP). Successful offline programming is defined as achieving $\{ x_\text{err} < 1\text{cm}, \; y_\text{err} < 1\text{cm}, \; \theta_\text{err} < 5^{\circ} \}$, where $x_\text{err}$, $y_\text{err}$, and $\theta_\text{err}$ are the differences between the final and target states. We list the statistical results in Table \ref{tab:offline performance}. The success rate of ZS-DDP is about 40\%, indicating many poor local minima that trap ZS-DDP in bad regions. Initializing with demonstrations increases the success rate by nearly 20\%, showing the benefits of using human demonstrations to avoid poor local minima. Our proposed demonstration-guided hierarchical framework, composed of DP-DDP and WS-DDP, achieves a significantly higher success rate (85\%) by using demonstrations as soft constraints and warm-starting a new DDP. We consider using demonstration as soft constraints, rather than initialization, as the key feature of our proposed method. This provides the nonlinear optimization process with a few but valuable human guidance, leading to a more efficient exploration process. WS-DDP is then warm-started by DP-DDP, yielding a much smoother and more precise optimal solution. In practice, as long as the three selected demonstrations are informative, the generalization results are not significantly impacted. It would be studied in the future about how to collect demonstrations more efficiently and actively.

\begin{figure}
	\begin{minipage}[b]{.48\linewidth}
		\centering
		\includegraphics[width=1\columnwidth]{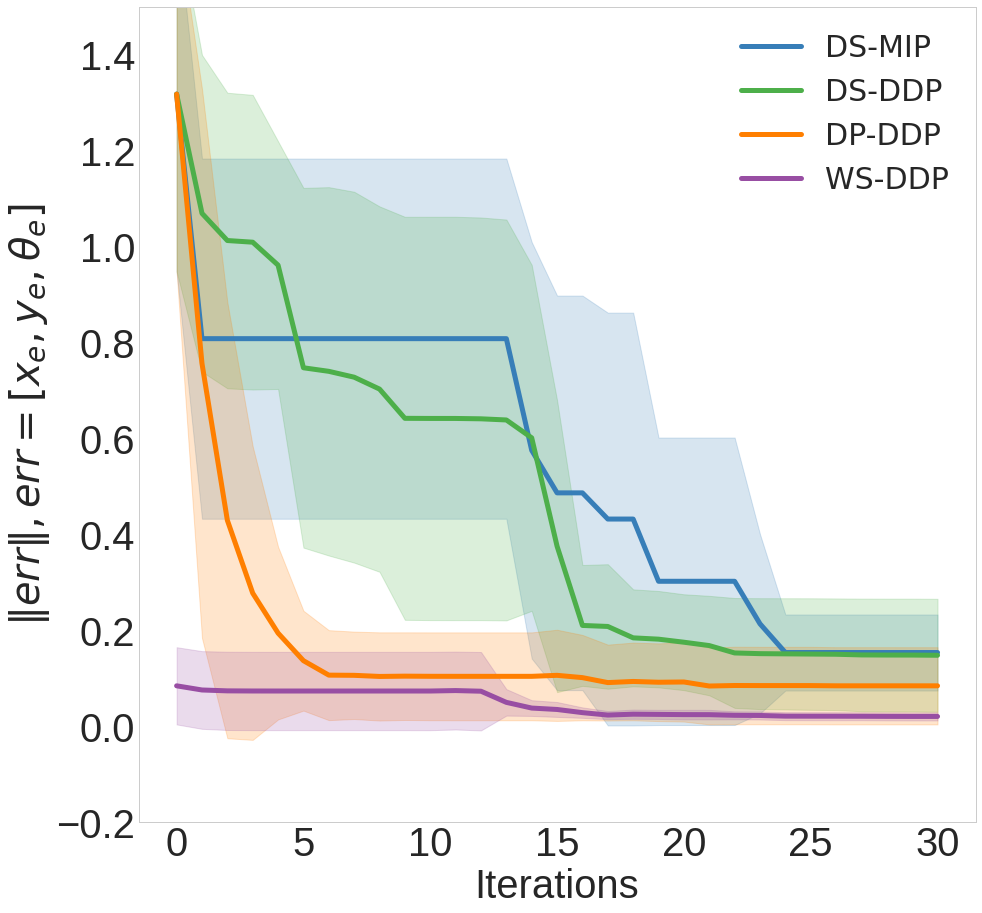}
		\caption{Convergence} 
		\label{fig:offline_comp}
	\end{minipage} 
	\hspace{-0.8cm}
	\hfill
	\begin{minipage}[b]{.52\linewidth}
		\centering
		\begin{tabular}{cc}
			\centering
			\includegraphics[width=0.45\linewidth]{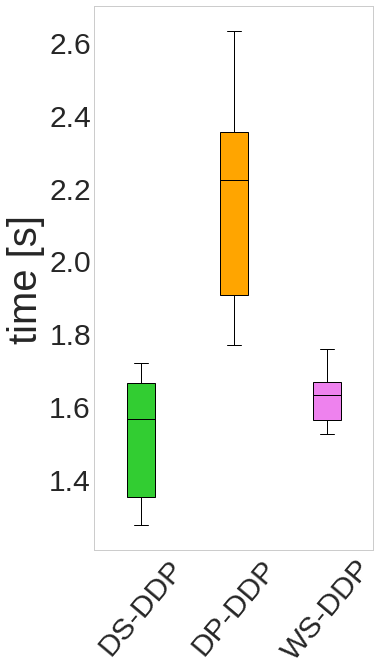} & 
			\hspace{-0.5cm}
			\includegraphics[width=0.44\linewidth]{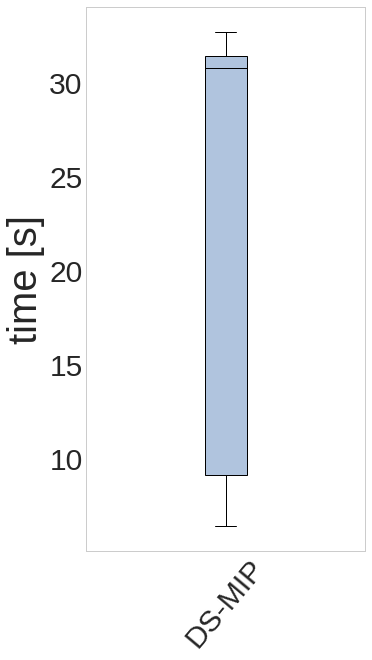}\\
		\end{tabular}
		\captionof{figure}{Computation time }  
		\label{fig:offline_time_comp}      
	\end{minipage}
\end{figure}

To demonstrate the strengths of our method, we moreover compared it with the Mixed-Integer Programming (MIP) formulation, which was commonly used for planar pushing tasks previously \cite{hogan_feedback_2020, de2022non, hogan2018reactive}. To ensure a fair comparison, we used demonstrations as initialization for MIP as well, which we refer to as DS-MIP. We used CasADi \cite{andersson2019casadi} with Bonmin solver \cite{bonami2012algorithms} and Crocoddyl \cite{mastalli2020crocoddyl} with FDDP solver to solve DS-MIP and DS-DDP, respectively. Convergence and computation time of 10 randomly selected target configurations are shown in Fig. \ref{fig:offline_comp} and Fig. \ref{fig:offline_time_comp}. The vertical axis of Fig. \ref{fig:offline_comp} represents the reaching error, in meters for $x_e$ and $y_e$ and in radians for $\theta_e$, between the final state $\bm{x}_T$ and the target $\bm{\mu}_T$. DS-MIP achieves almost the same result as DS-DDP, as shown in Fig. \ref{fig:offline_comp}, but takes ten times longer to solve, as illustrated in Fig. \ref{fig:offline_time_comp}. Moreover, our experiments demonstrate that the proposed DP-DDP and WS-DDP algorithms converge to better results, albeit over relatively longer time (still within acceptable limits), in comparison to DS-DDP. DS-DDP was found to converge to poor local optima and increasing the number of iterations did not improve the results. These findings suggest that initializing solely using demonstrations is insufficient, but constraining the search space with demonstrations as soft constraints can lead to better solutions.

\renewcommand{\arraystretch}{1.}
\begin{table}[t]
	\caption{Performance of ZS-DDP, DS-DDP, DP-DDP, and WS-DDP for offline programming}
	\begin{footnotesize}
		\begin{tabular}{l | c c c c}
			\toprule
			{\bf{Method}}  & $\bm{x_\text{err}}$ (cm) & $\bm{y_\text{err}}$ (cm) & $\bm{\theta_\text{err}}$ (rad) & $\text{succ\_rate}$ \\
			\midrule
                {ZS-DDP} & 3.18  $\pm$ 8.30  &  4.78  $\pm$  8.60 & 0.04  $\pm$ 0.39 & 40\%\\
			{DS-DDP} & 1.70  $\pm$ 9.07  &  1.62  $\pm$  9.66 & 0.21  $\pm$ 1.82 & 58\%\\
			{DP-DDP} & 0.14  $\pm$ 1.54  &  0.57  $\pm$  2.77 & 0.22  $\pm$ 0.12 & 77\%\\
			{\textbf{WS-DDP}} & \textbf{0.07  $\pm$ 1.31}  &  \textbf{0.19  $\pm$  2.90} & \textbf{0.01  $\pm$ 0.08} & \textbf{85\%} \\
			\bottomrule
		\end{tabular}
	\end{footnotesize}
	\label{tab:offline performance}
\end{table}

\begin{figure}[t!]
	\centering
	\includegraphics[width=0.7\columnwidth]{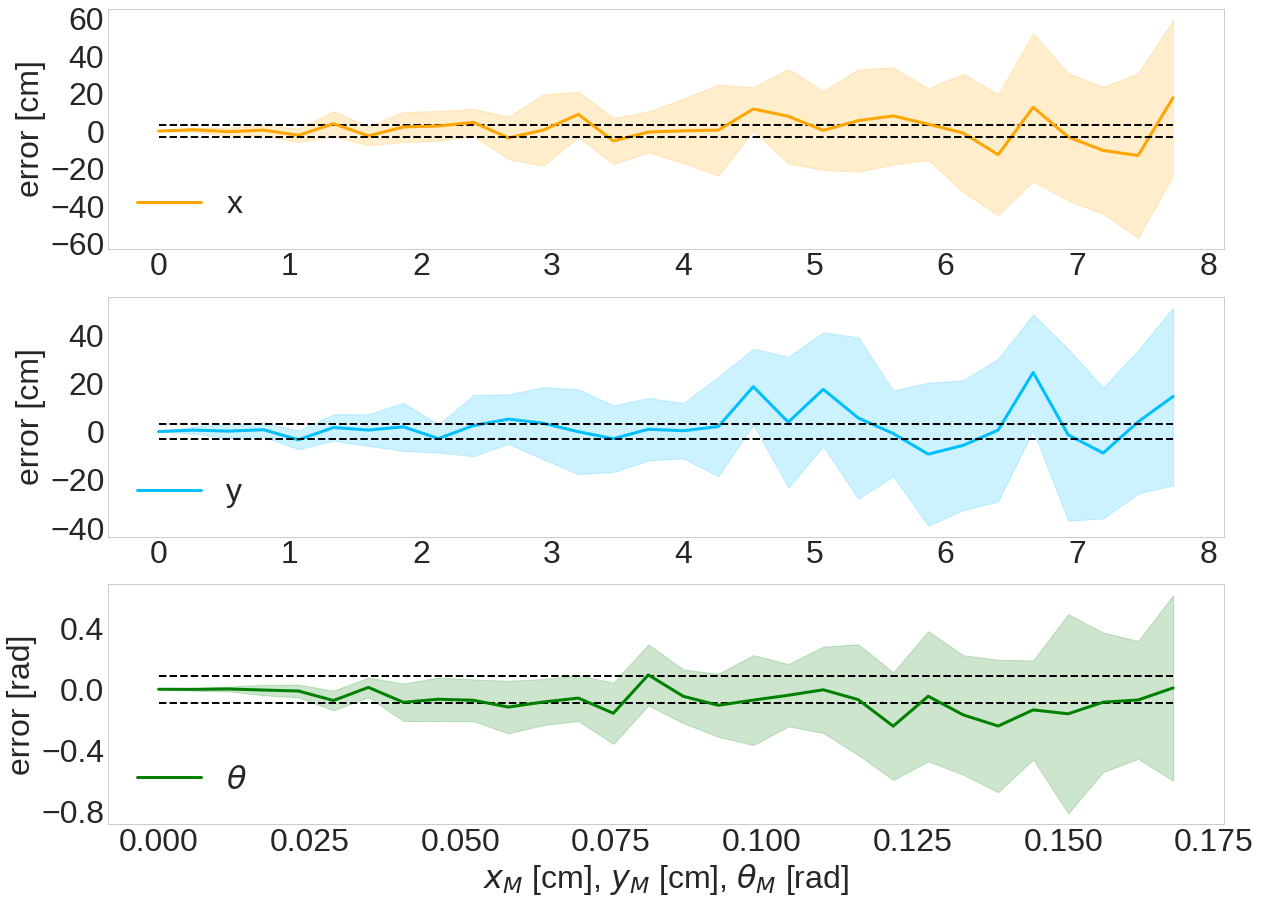}
	\caption{Tracking performance under disturbances drawn from uniform distributions, where $x_M$, $y_M$ and $\theta_M$ are the boundaries. The dashed lines present the tolerance.}
	\label{fig:dis_perf}
	\vspace{-0.5cm}
\end{figure}

\begin{figure*}[htbp] 
	\centering
	\subfloat[Initialization]{{\includegraphics[width=0.3\columnwidth]{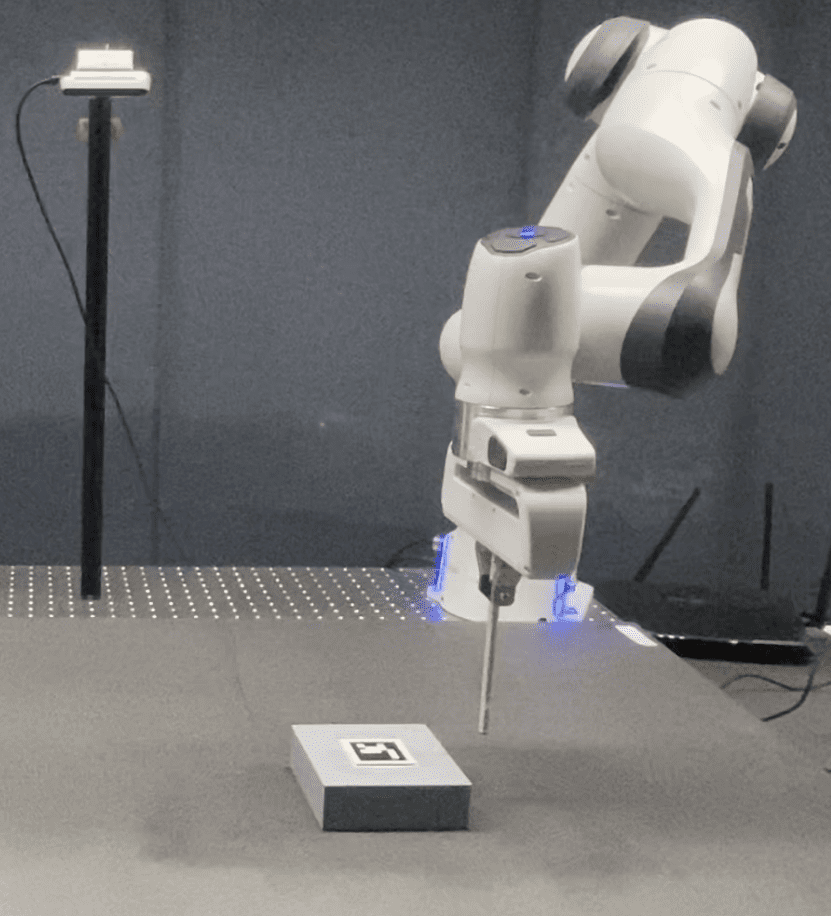}}}
	\subfloat[Contact]{{\includegraphics[width=0.3\columnwidth]{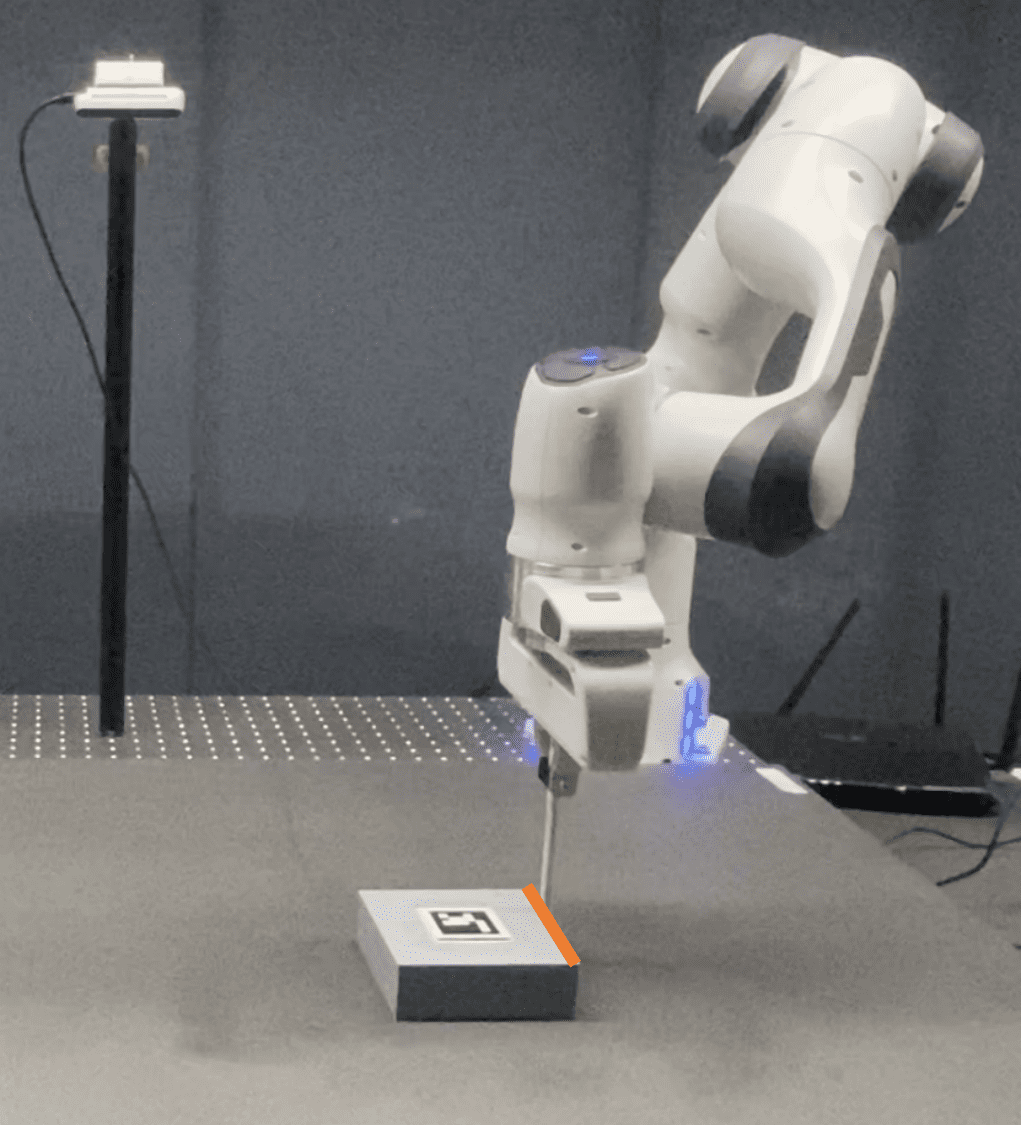}}} 
	\subfloat[Pushing]{{\includegraphics[width=0.3\columnwidth]{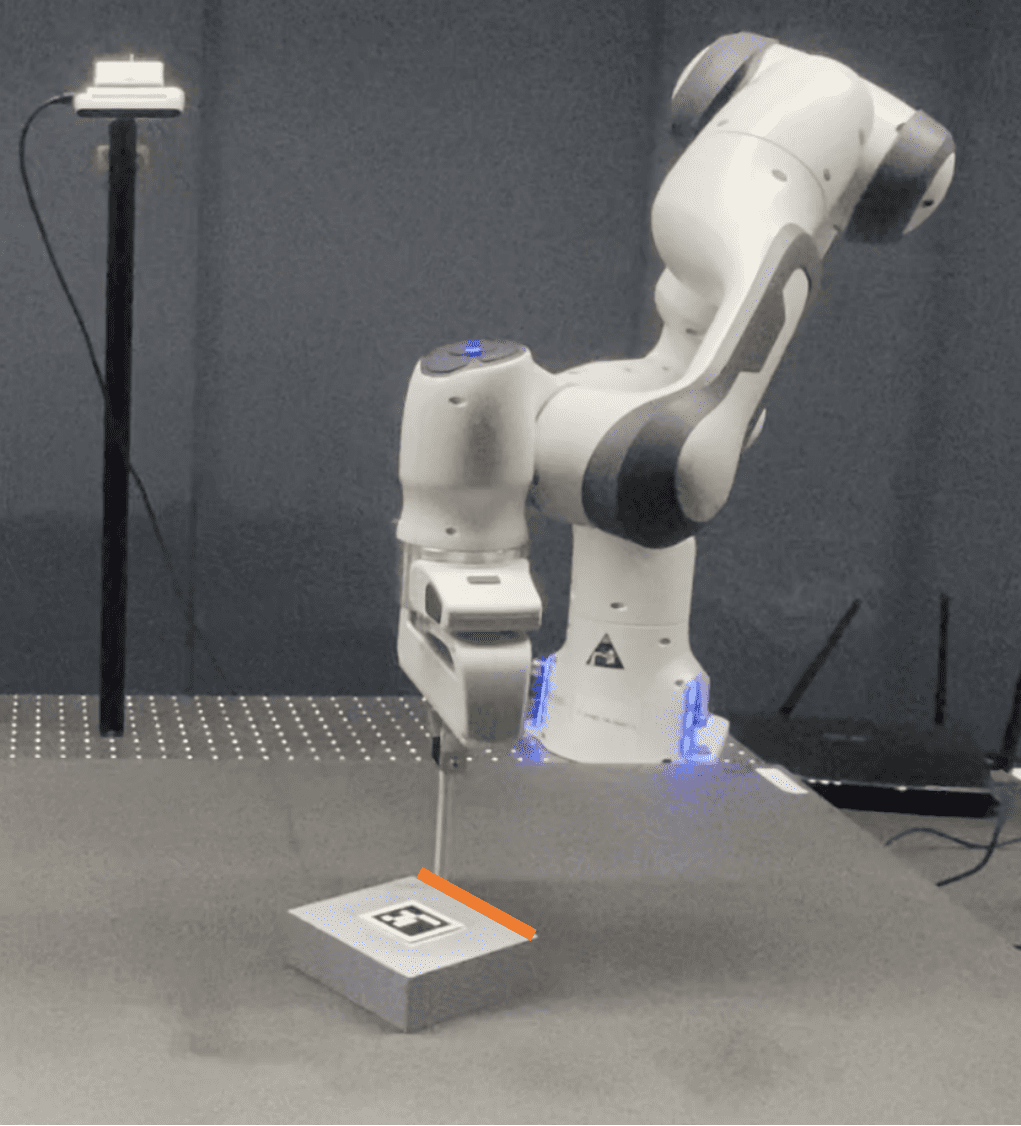}}} 
	\subfloat[Face switching]{{\includegraphics[width=0.3\columnwidth]{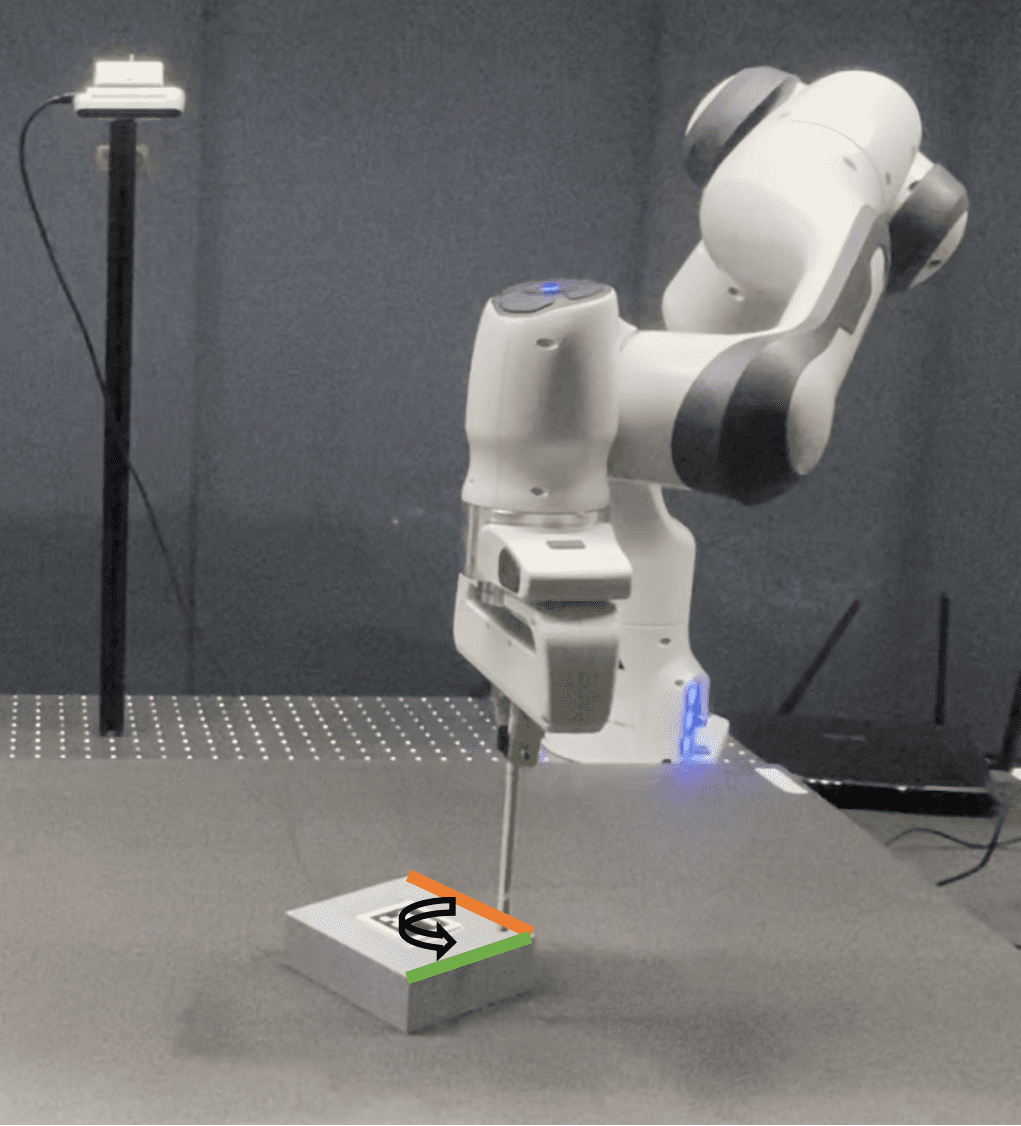}}} 
	\subfloat[Contact]{{\includegraphics[width=0.3\columnwidth]{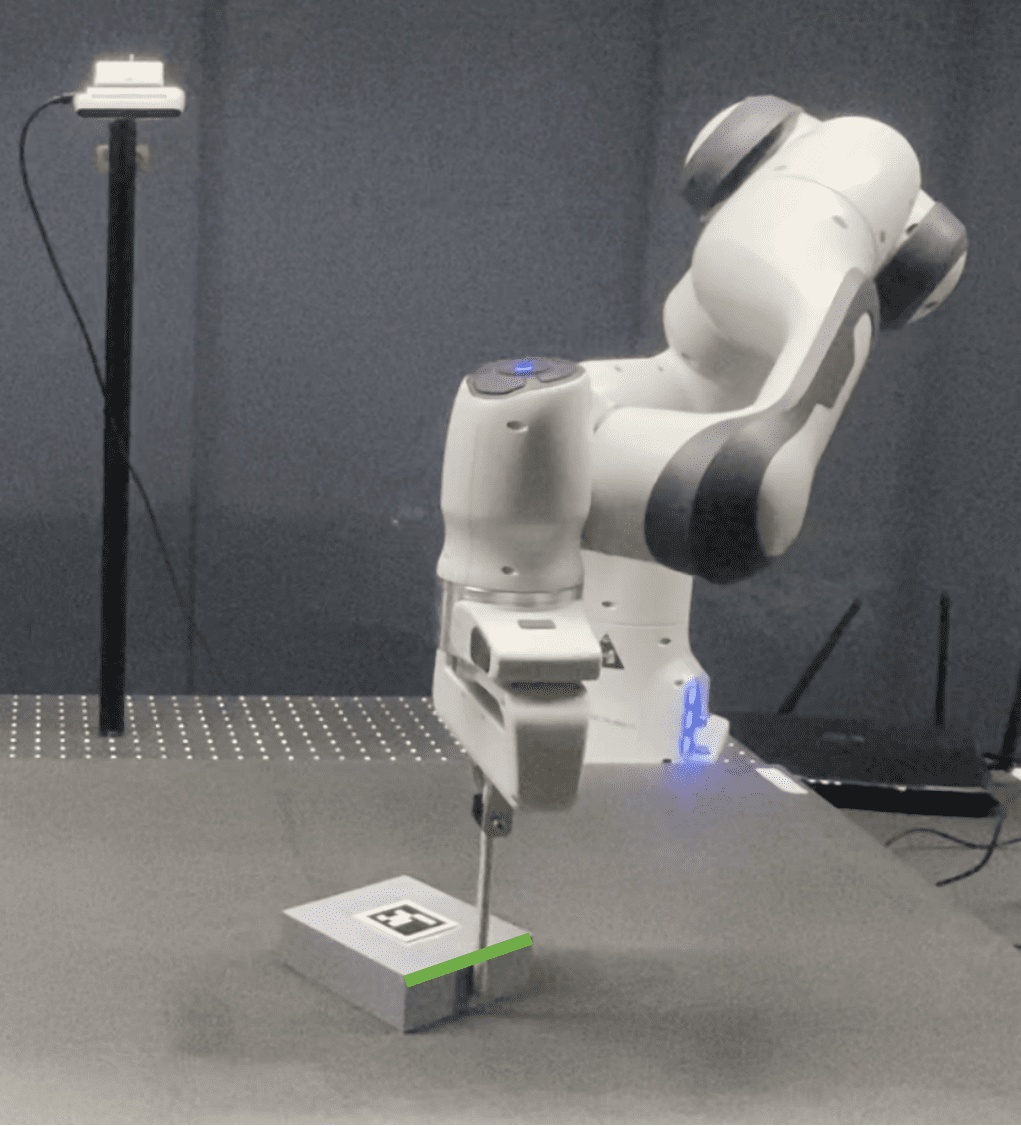}}} 
	\subfloat[Pushing]{{\includegraphics[width=0.3\columnwidth]{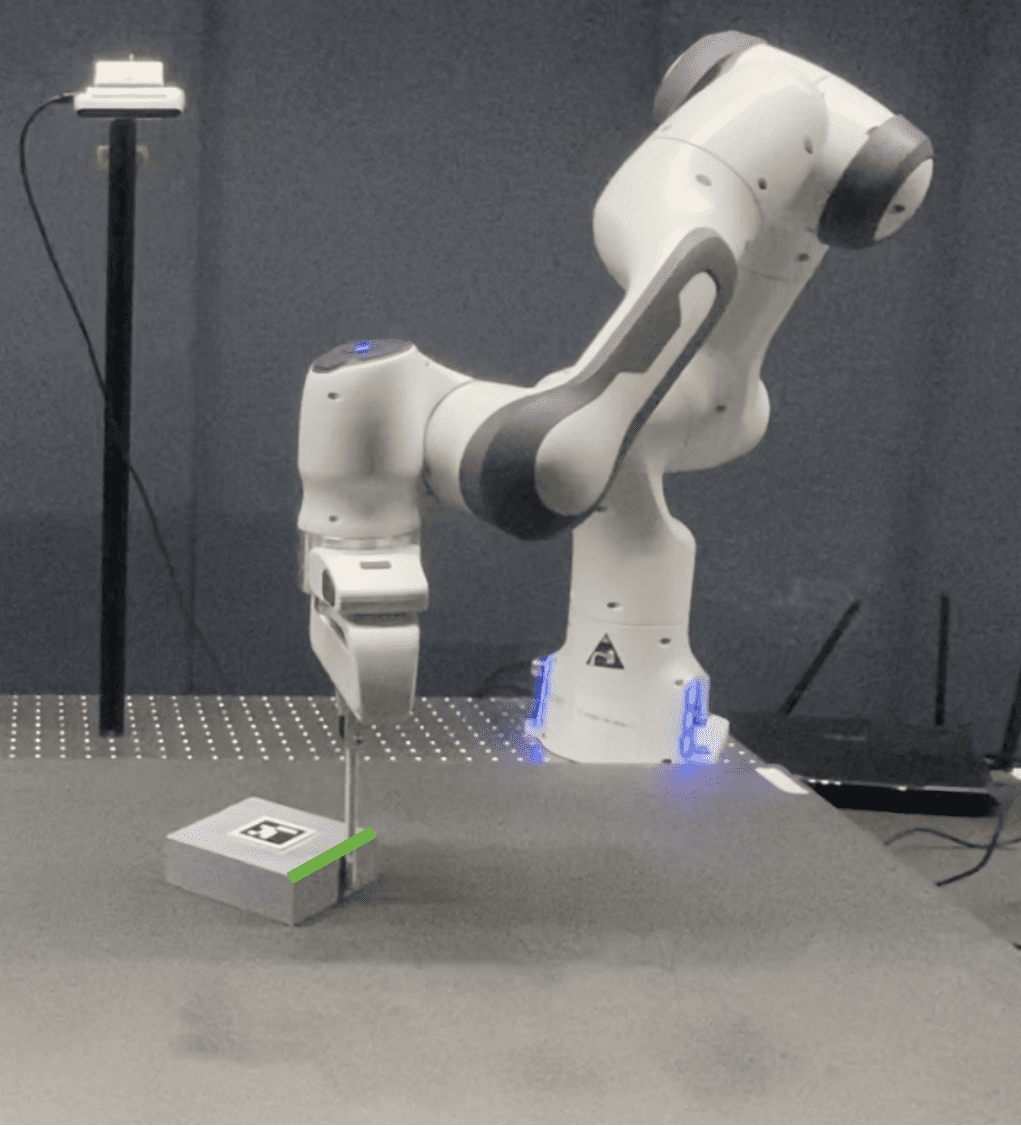}}} 
	\subfloat[Reaching]{{\includegraphics[width=0.3\columnwidth]{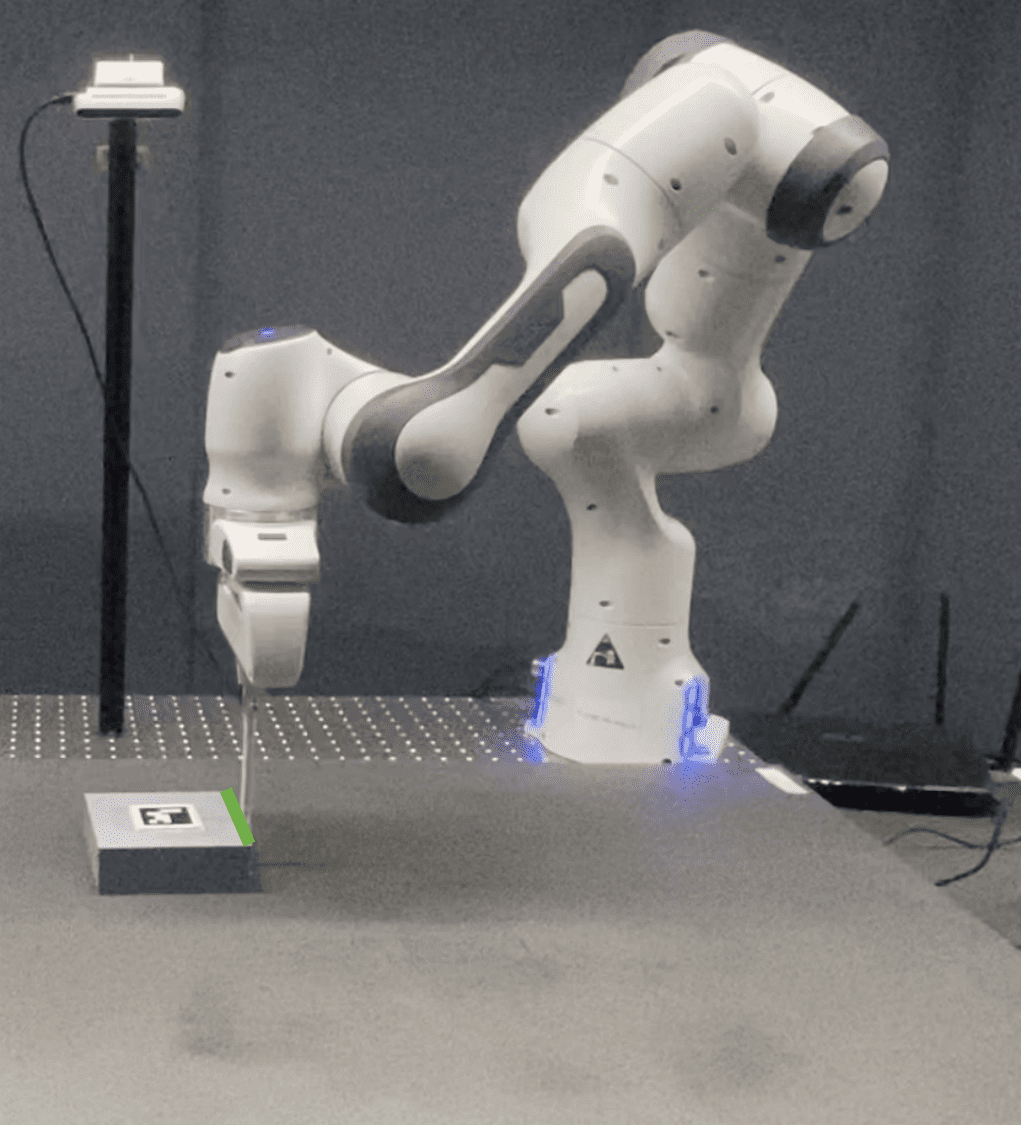}}}
	\caption{Pushing task with face switching. The manipulator starts from (a) and selects an optimal face (orange) and contact point (b) for pushing, until reaching the planned face-switching point (c). Next, the manipulator changes to another face (d) and touches the object again at the selected point (e), followed by the next phase of pushing (f), until reaching the final target (g). This example is for $N_s=1$. (d)$\sim$(f) should be repeated if $N_s>1$. The colored lines are used to express the current active faces, and the black arrow in (d) represents the face-switching process.}
	\label{fig:keyframe}
	\vspace{-0.5cm}
\end{figure*}


\subsection{Online Tracking}
\label{sec:ex_online tracking}

For online tracking, we investigated both numerical simulation and real robot experiments. In simulation, we introduced a disturbance on the state as $\bm{x} = \overline{\bm{x}} + \bm{\epsilon}$ from the beginning to the end, where $\epsilon_x \sim \mathbb{U}(-x_M, x_M)$, $\epsilon_y \sim \mathbb{U}(-y_M, y_M)$, $\epsilon_\theta \sim \mathbb{U}(-\theta_M, \theta_M)$, are the components of $\bm{\epsilon}$, drawn from a uniform distributions. Fig. \ref{fig:dis_perf} shows the evolution of the errors on $x$, $y$, and $\theta$, computed as the difference between the final point and the target, for increasing $x_M$, $y_M$ and $\theta_M$. The tolerance for online tracking is set as $\{ x_\text{err} < 3\text{cm},\; y_\text{err} < 3\text{cm},\;  \theta_\text{err} < 5^{\circ} (0.087\text{rad}) \}$. The controller can successfully resist 4cm perturbation for $x$ and $y$, and 0.117rad for $\theta$.

\begin{table}[t!]
	\setlength\tabcolsep{0.005pt}
	\begin{center}
		\begin{tabular}{cc}
			\centering
			\includegraphics[width=0.45\linewidth]{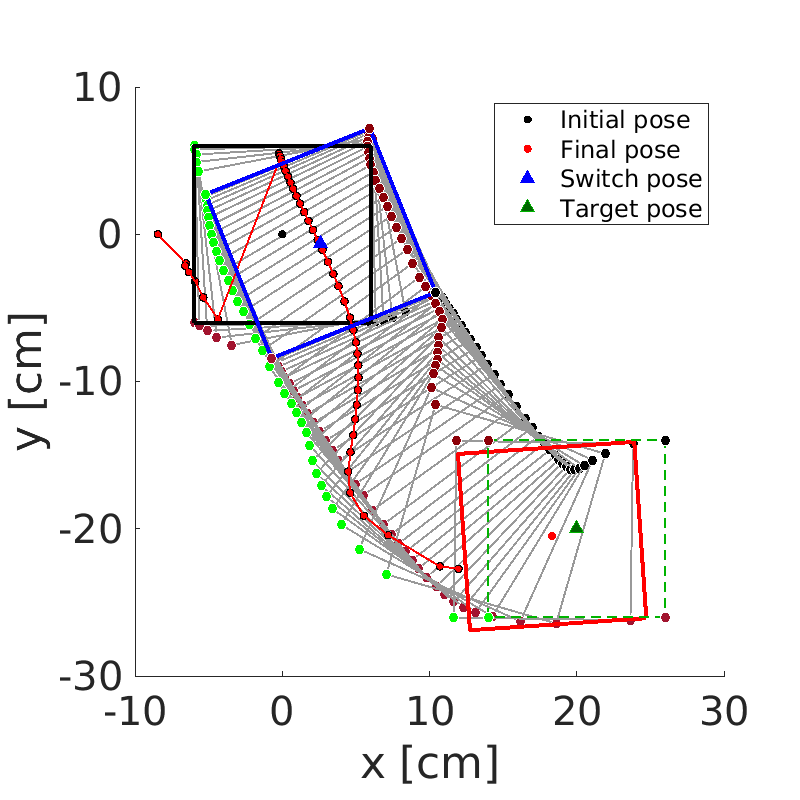} & 
			\includegraphics[width=0.45\linewidth]{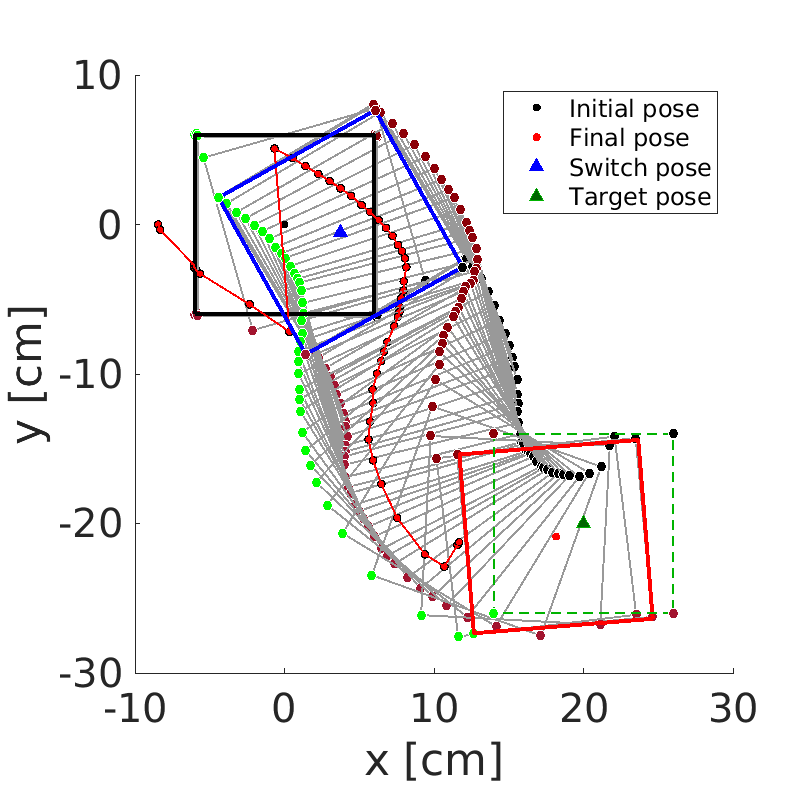} \\ 
			a) Simulation results & b) Experimental results\\
		\end{tabular}
		\captionof{figure}{Planar pushing with face switching. Both simulation and experimental results reach final targets within tolerance.}    
		\label{fig:real_comp}    
	\end{center}
	\vspace{-1cm}
\end{table}

Then, we tested the proposed method on the real robot setup (Fig. \ref{fig:real_setup}), using a 7-axis Franka Emika robot and a RealSense D435 camera. The prismatic slider, a 12$\times$12cm block made of 3D-printed PLA and weighting 110g, was positioned on a flat plywood surface, with an Aruco Marker attached to its top face. The friction coefficients between the pusher and the slider, and between the plywood surface and the slider, are 0.3 and 0.35, respectively. The robot was equipped with a wooden pusher that had a 0.5cm radius and was used to move the slider. The camera captured the motion of the slider at 30Hz, while the feedback controller ran at 100Hz. The low-level Cartesian impedance controller, responsible for actuating the robot, ran at 1000Hz.

As reported in \cite{doshi2020hybrid}, most targets can be achieved with only one to two face switches. We therefore tested our approach with one externally disturbed line tracking task and two face-switching tasks (see accompanying video). The trajectories in Fig. \ref{fig:real_comp} correspond to the task requiring one face switching to push the object from $[0, 0, 0]$ to $[20\text{cm}, \minus20\text{cm}, \pi/2]$. The control strategy is intuitive: pushing it from the left face to slightly adjust the pose and then changing to the top face for the next phase of pushing. Fig. \ref{fig:real_comp}-(a) is the simulation trajectory, which is generated by using PyBullet \cite{coumans2016pybullet}, while Fig. \ref{fig:real_comp}-(b) shows the real robot trajectory. It is observed that these two trajectories are significantly different, because of the different friction parameters in the two different worlds. Still, both can overcome the uncertainty to reach the final target. Despite the existence of unstructured elements such as differences in the visual system and the robot controller, several assumptions of the dynamics model, as well as immeasurable friction, the feedback controller is able to cope with these different mismatches and track the reference trajectory successfully. Fig. \ref{fig:keyframe} shows the keyframes of the pusher pushing the slider toward the target, indicating that 7 steps are needed with the face-switching strategy. Another final target configuration, $[5\text{cm}, \minus18\text{cm}, \pi/5]$, which requires two face switches, is additionally shown in the video.

%% file: sections/conclusion.tex
\section{Discussion, Limitations and Conclusion}


In this paper, we proposed to add separation mode and face-switching mechanism to the problem of pushing objects on a planar surface. We showed that by introducing only few human demonstrations, the typical TAMP problem can be expressed as a classical continuous optimization problem, which is much more efficient to be solved. With the proposed demonstration-guided hierarchical optimization framework, we demonstrated significantly better results in terms of generalization and precision compared to the state-of-the-art methods. Additionally, we developed a feedback controller based on DDP feedback gains to regenerate the trajectory for online tracking. We tested the combination of these approaches in both PyBullet and real robot applications, showing good performance to resist contact uncertainty. 

Currently, we are using a feedback controller to track the offline trajectory. If the system is subject to large perturbations such as rotating $180^\circ$, an online Model Predictive Control (MPC) may still be required. Nevertheless, our demonstration-guided method is also promising as an optimizer within MPC to avoid poor local optima.

As future work, we aim to apply our demonstration-guided approach to a broader range of manipulation tasks (namely, beyond pushing problems), which requires further study on how to extract and represent constraints from demonstrations. It would also be relevant to explore extensions of TAMP problems by introducing similar continuous human demonstrations. Further work could also exploit human demonstrations in model-based learning strategies to let the robot automatically refine the pushing model and its motion.